\documentclass[twoside,11pt]{article}
\usepackage{jair, theapa, rawfonts}
\usepackage{multirow}
\usepackage{graphicx}
\usepackage{color}
\usepackage{url, xspace, textcomp}
\usepackage{amssymb,amsmath,amsfonts}
\usepackage[titletoc,toc]{appendix}
\usepackage{subcaption}
 \usepackage[update,prepend]{epstopdf}
\usepackage{algorithm}
\usepackage{algorithmic}
\usepackage{xfunctions}
\usepackage{arydshln}

\newcommand{\fb}{\textsc{FB15k}\xspace}
\newcommand{\svo}{\textsc{SVO}\xspace}
\newcommand{\kinships}{\textsc{Kinships}\xspace}
\newcommand{\umls}{\textsc{UMLS}\xspace}

\newcommand{\su}{{\it head}\xspace}
\newcommand{\ve}{{\it label}\xspace}
\newcommand{\ob}{{\it tail}\xspace}

\newcommand{\bi}{\textsc{Bigrams}\xspace}
\newcommand{\tri}{\textsc{Trigram}\xspace}
\newcommand{\tatec}{\textsc{Tatec}\xspace}
\newcommand{\tateclc}{\textsc{Tatec-lc}\xspace}
\newcommand{\tatecft}{\textsc{Tatec-ft}\xspace}

\newcommand{\transe}{\textsc{TransE}\xspace}
\newcommand{\rescal}{\textsc{RESCAL}\xspace}
\newcommand{\transh}{\textsc{TransH}\xspace}
\newcommand{\transr}{\textsc{TransR}\xspace}
\newcommand{\ctransr}{\textsc{cTransR}\xspace}

\newcommand{\smel}{{\sf SME}(linear)\xspace}
\newcommand{\smebl}{{\sf SME}(bilinear)\xspace}
\newcommand{\lfm}{{\sf LFM}\xspace}
\newcommand{\ntn}{{\sf NTN}\xspace}

\newcommand{\soft}{{-\scriptsize soft}\xspace}
\newcommand{\hard}{{-\scriptsize hard}\xspace}

\newcommand{\citebi}{{\bf (B)}\xspace}
\newcommand{\citetri}{{\bf (T)}\xspace}

\newcommand{\norm}[1]{{\parallel#1\parallel}_2}
\newcommand{\hd}{h}
\newcommand{\tl}{t}

\newcommand{\lbl}{\ell}
\newcommand{\ents}{{\cal E}}
\newcommand{\labs}{{\cal L}}

\newcommand{\nents}{E}
\newcommand{\nlabs}{L}
\newcommand{\ent}{e}
\newcommand{\lab}{l}
\newcommand{\intint}[1]{[\![#1]\!]}
\renewcommand{\Re}{\mathbb{R}}
\newcommand{\trp}{(\hd,\lbl, \tl)}
\newcommand{\cortrp}{{\cal C}\trp}
\newcommand{\hdprime}{\hd'}
\newcommand{\lblprime}{\lbl'}
\newcommand{\tlprime}{\tl'}
\newcommand{\trpprime}{(\hdprime, \lblprime,\tlprime)}
\newcommand{\trpprimee}{(\hdprime, \lbl,\tlprime)}
\newcommand{\trpprimeee}{(\hd, \lblprime,\tl)}
\newcommand{\sco}[1]{s#1}
\newcommand{\biterm}{1}
\newcommand{\triterm}{2}
\newcommand{\bitermhead}{1}
\newcommand{\bitermtail}{2}
\newcommand{\scorb}[1]{s_\biterm#1} 
\newcommand{\scort}[1]{s_\triterm#1} 
\newcommand{\eh}{{\bf e}^\hd}
\newcommand{\et}{{\bf e}^\tl}
\newcommand{\rel}{{\bf r}^\ell} 
\newcommand{\w}{{\bf w}^\ell}
 
\newcommand{\ehb}{{\bf e}^\hd_\biterm} 
\newcommand{\eht}{{\bf e}^\hd_\triterm} 
\newcommand{\etb}{{\bf e}^\tl_\biterm} 
\newcommand{\ett}{{\bf e}^\tl_\triterm} 
\newcommand{\rhb}{{\bf r}^\ell_{\bitermhead}} 
\newcommand{\rtb}{{\bf r}^\ell_{\bitermtail}} 
\newcommand{\rt}{{\bf R}^\ell} 
\newcommand{\diago}{{\bf D}}
\newcommand{\calS}{{\cal S}}
\newcommand{\dotb}[2]{\big<#1\big|#2\big>}
\newcommand{\dott}[3]{\big<#1\big|#2\big|#3\big>}
\newcommand{\dbi}{{d_\biterm}}
\newcommand{\dtri}{{d_\triterm}}

\jairheading{1}{2015}{1-15}{6/91}{9/91}
\ShortHeadings{ Embeddings Models for Link Prediction in KBs}
{Garc\'ia-Dur\'an, Bordes, Usunier \& Grandvalet}
\firstpageno{1}

\begin{document}

\title{Combining Two And Three-Way Embeddings Models for Link Prediction in Knowledge Bases}

\author{\name Alberto Garc\'ia-Dur\'an \email alberto.garcia-duran@utc.fr \\
  \addr Sorbonne universit\'es, Universit\'e de technologie de
  Compi\`egne, CNRS,  Heudiasyc UMR 7253\\
 CS 60 319, 60 203 Compi\`egne cedex, France
  \AND 
  \name Antoine Bordes \email abordes@fb.com \\
  \addr Facebook AI Research\\
  770 Broadway, New York, NY 10003. USA
\AND
  \name Nicolas Usunier\thanks{Part of this work was
    done while Nicolas Usunier was with Sorbonne universit\'es, Universit\'e de technologie de
  Compi\`egne, CNRS,  Heudiasyc UMR 7253.} \email usunier@fb.com\\
  \addr Facebook AI Research\\
  112 Avenue de Wagram, 75017 Paris, France
  \AND
  \name Yves Grandvalet \email yves.grandvalet@utc.fr \\
  \addr Sorbonne universit\'es, Universit\'e de technologie de
  Compi\`egne, CNRS,  Heudiasyc UMR 7253\\
 CS 60 319, 60 203 Compi\`egne cedex, France
	}


\maketitle

\begin{abstract}
  This paper tackles the problem of endogenous link prediction for
  Knowledge Base completion. Knowledge Bases can be represented as
  directed graphs whose nodes correspond to entities and edges to
  relationships.
  Previous attempts either consist of powerful systems with high
  capacity to model complex connectivity patterns, which unfortunately
  usually end up overfitting on rare relationships, or in approaches
  that trade capacity for simplicity in order to fairly model all
  relationships, frequent or not. In this paper, we propose \tatec a
  happy medium obtained by complementing a high-capacity model with a
  simpler one, both pre-trained separately and then combined.
  We present several variants of this model with different kinds of
  regularization and combination strategies and show that this approach
  outperforms existing methods on different types of relationships by
  achieving state-of-the-art results on four benchmarks of the literature.
\end{abstract}

\section{Introduction}
\label{Introduction}
Knowledge bases (KBs) are crucial tools to deal with the rise of data, since they provide ways to organize, manage and retrieve all digital knowledge. These repositories can cover any kind of area, from specific domains like biological processes (e.g. in {\sc GeneOntology}\footnote{\url{http://geneontology.org}}), to very generic purposes. {\sc Freebase}\footnote{\url{http://www.freebase.com}}, a huge collaborative KB which belongs to the Google Knowledge Graph, is an example of the latter kind which provides expert/common-level knowledge and capabilities to its users.
 An example of knowledge engine is {\sc WolframAlpha}\footnote{\url{http://www.wolframalpha.com}}, an engine which answers to any natural language question, like \texttt{how far is saturn from the sun?}, with human-readable answers (\texttt{1,492 $\times 10^9$ km}) using an internal KB. Such KBs can be used for question answering, but also for other natural language processing tasks like word-sense disambiguation \cite{navigli2005structural}, co-reference resolution \cite{ponzetto2006exploiting} or even machine translation \cite{knight1994building}.

KBs can be formalized as directed multi-relational graphs, whose nodes correspond to entities connected with edges encoding various kinds of relationship. Hence, one can also refer to them as multi-relational data. In the following we denote connections among entities via triples or {\it facts} (\textit{head, label, tail}), where the entities \textit{head} and \textit{tail} are connected by the relationship \textit{label}. Any information of the KB can be represented via a triple or a concatenation of several ones. Note that multi-relational data are not only present in KBs but also in recommender systems, where the nodes would correspond to users and products and edges to different relationships between them, or in social networks for instance.

\begin{figure}[t]
\begin{center}
\vspace*{-1ex}
\includegraphics[width=0.5\linewidth]{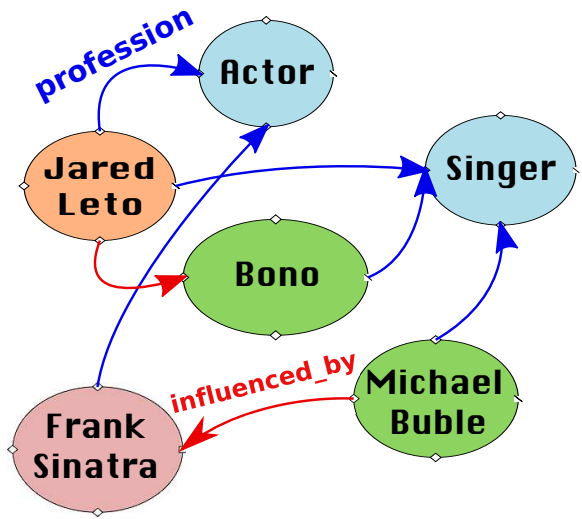}
\caption{\label{fig:ex} {\bf Example of (incomplete) Knowledge Base} with
  6 entities, 2 relationships and 7 facts.} 
\end{center}
\vspace*{-5ex}
\end{figure}

A main issue with KBs is that they are far from being complete. {\sc Freebase} currently contains thousands of relationships and more than 80 millions of entities, leading to billions of facts, but this remains only a very small portion out of all the human knowledge, obviously. And since question answering engines based on KBs like  {\sc WolframAlpha} are not capable of generalizing over their acquired knowledge to fill in for missing facts, they are {\it de facto} limited: they search for matches with a question/query in their internal KB and if this information is missing they can not provide a correct answer, even if they correctly interpreted the question.
Consequently, huge efforts are devoted nowadays towards KBs construction or completion, via manual or automatic processes, or a mix of both. This is mainly divided in two tasks: entity creation or extraction, which consists in adding new entities to the KB and link prediction, which attempts to add connections between entities. This paper focuses on the latter case.
Performing link prediction can be formalized as filling in incomplete
triples like ({\it head}, {\it label}, ?) or (?, {\it label}, {\it
  tail}), by predicting the missing argument of the triple when such
triple  does not exist in the KB, yet. For instance, given the small
example KB of Figure~\ref{fig:ex}, made of 6 entities and 2 different
relationships, and containing facts like (\texttt{Jared Leto,
  influenced\_by, Bono}) or (\texttt{Michael Buble, profession,
  singer}), we would like to able to predict new links such as
(\texttt{Frank Sinatra, profession, singer}), by using the fact
that he influenced the singer \texttt{Michael Buble} for instance. 

Link prediction in KBs is complex due to several issues. The entities are not homogeneously connected: some of them will have a lot of links with other entities, whereas others will be rarely connected. To illustrate the diverse characteristics present in the relationships we can take a look at \fb, a subset of {\sc Freebase} introduced in \cite{bordesNIPS13}. In this data set of $\sim$14k entities and 1k types of relationships, entities have a mean number of triples  of $\sim$400, but a median of 21 indicating that a large number of them appear in very few triples. Besides, roughly 25$\%$ of connections are of type 1-to-1, that is, a head is connected to at most one tail, and around 25$\%$ are of type Many-to-Many, that is, multiple heads can be linked to a tail and vice versa. 
As a result, diverse problems coexist in the same database.
Another property of relationships that can have a big impact on the performance is the typing of their arguments. On \fb, some relationships are very strongly typed like \texttt{/sports/sports$\_$team/location}, where one always expects a football team as head and a location as tail, and some are far less precise such as \texttt{/common/webpage/category} where one expects only web page adresses as tail but pretty much everything else as head.
A link prediction algorithm should be able to adapt to these different settings. 

Though there exists (pseudo-) symbolic approaches for link prediction
based on Markov-logic networks \cite{Kok:2007} or random walks
\cite{lao2011random}, learning latent features representations of KBs
constituents - the so-called {\it embedding methods} - have recently
proved to be more efficient for performing link prediction in KBs,
e.g. in
\cite{bordesNIPS13,wang2014knowledge,lin2015learning,chang2014typed,wangknowledge,zhangword,yang2014joint}. In
all these works, entities are represented by low-dimensional vectors -
the embeddings - and relationships act as operators on them: both
embeddings and operators define a scoring function that is learned so
that triples observed in the KBs have higher scores than unobserved
ones. The embeddings are meant to capture underlying features that
should eventually allow to create new links successfully. The scoring
function is used to predict new links: the higher the score, the more
likely a triple is to be true. Representations of relationships are
usually specific (except in \lfm \cite{jenatton2012latent} where there
is a sharing of parameters across relationships), but embeddings of
entities are shared for all relationships and allow to transfer information across them. The learning process can be considered as multi-task, where one task concerns each relationship, and entities are shared across tasks.

Embedding models can be classified according to the interactions that they use to encode the validity of a triple in their scoring function. If the joint interaction between the head, the label and the tail is used then we are dealing with a {\it 3-way} model; but when the binary interactions between the head and the tail, the head and the label, and the label and the tail are the core of the model, then it is a {\it 2-way} model. Both kinds of models represent the entities as vectors, but they differ in the way they model the relationships: 3-way models generally use matrices, whereas 2-way models use vectors. This difference in the capacity leads to a difference in the expressiveness of the models. The larger capacity of 3-way models (due to the large number of free parameters in matrices) is beneficial for the relationships appearing in a lot of triples, but detrimental for rare ones even if regularization is applied. Capacity is not the only difference between 2- and 3-way models, the information encoded by these two models is also different: we show in Sections \ref{TATEC} and \ref{anecdotal} that both kinds of models assess the validity of the triple using  different data patterns.

In this paper we introduce \tatec that encompass previous works by combining well-controlled 2-way interactions with high-capacity
3-way ones. We aim at capturing data patterns of both approaches by
separately pre-training the embeddings of 2-way and 3-way models and
using different embedding spaces for each of the two of them. We demonstrate in
the following that otherwise -- with no pre-training and/or no use of
different embedding spaces -- some features cannot be conveniently
captured by the embeddings. Eventually, these pre-trained weights are
combined in a second stage, leading to a combination model which
outperforms most previous works in all conditions on four benchmarks
from the literature, \umls, \kinships, \fb and \svo.
\tatec is also carefully regularized since we systematically compared two different regularization schemes: adding penalty terms to the loss function or hard-normalizing the embedding vectors by constraining their norms.

This paper is an extension of \cite{duranECML}: we added a much more thorough study on regularization and on combination strategies for \tatec. Besides we propose experiments on several new benchmarks and a more complete comparison of our proposed method w.r.t. the state-of-the-art. We also give examples of predictions and projections in 2D of the obtained embeddings to provide some insights into the behavior of \tatec. 
The paper is organized as follows. Section \ref{related} discusses
previous works. Section \ref{TATEC} presents our model and justifies
its choices. Detailed explanations of both the training
procedure and the regularization schemes are given in Section
\ref{train}. Finally, we present our experimental results on four
benchmarks in Section \ref{results}.

\section{Related work}
\label{related}

In this section, we discuss the state-of-the-art of modeling large
multi-relational databases, with a particular focus on embedding
methods for knowledge base completion.

One of the simplest and most successful 2-way models is \transe
\cite{bordesNIPS13}. In that model, relationships are represented as
translations in the embedding space: if $\trp$ holds, then the
embedding of the \ob $\tl$ should be close to the embedding of \su
$\hd$ plus some
vector that depends on the \ve $\lbl$. This is a natural approach to model
hierarchical and asymmetric relationships, which are common in
knowledge bases such as {\sc Freebase}. Several modifications to
\transe have been proposed recently, \transh \cite{wang2014knowledge}
and \transr \cite{lin2015learning}. In \transh, the embeddings of the
entities are projected onto a hyperplane that depends on $\lbl$ before
the translation. The second algorithm, \transr, follows the same idea,
except that the projection operator is a matrix that is more
general than an orthogonal projection to a hyperplane. As we shall see
in the next section, \transe corresponds to our \bi model with
additional constraints on the parameters.

While 2-way models were shown to have very good performances on some
KB datasets, they have limited expressiveness and they can fail
dramatically on harder datasets. In contrast, 3-way models perform
some form of low-rank tensor factorization, and in that respect can
have extremely high expressiveness depending on the rank
constraints. In the context of link prediction for multi-relational
data, \rescal \cite{Nickel:2011} follows natural modeling
assumptions. Similarly to \transe, \rescal learns one low-dimensional
embedding for each entity. However, relationships are represented as a
bilinear operator in the embedding space, i.e. each relationship
corresponds to a matrix. The training objective of \rescal is the
Frobenius norm between the original data tensor and its low-rank
reconstruction, whereas \tatec uses the margin ranking criterion of
\transe. Another related 3-way model is \smebl \cite{bordesMLJ13}. The
parameterization of \smebl is a constrained version of \rescal, and
also uses a ranking criterion as training objective.  

The Latent Factor Model (\lfm) \cite{jenatton2012latent} and the
Neural Tensor Networks (\ntn) \cite{SocherChenManningNg2013} use
combinations of a 3-way model with a more constrained 2-way model, and
in that sense are closer to our algorithm \tatec. There are important
differences between these algorithms and \tatec, though.
First, both \lfm and \ntn share the entity embeddings in the 2-way and
the 3-way models, while we learn different entity embeddings. The use
of different embeddings for the 2-way and the 3-way models does not
increase the model expressivity, because it is equivalent to a
combination with shared embeddings in a higher dimensional embedding
space, with additional constrains on the relation parameters. As we
show in the experiments however, these additional constraints lead to
very significant improvements.
The second main difference between our approach and \lfm is that some
parameters of the relationships between the 2-way and the 3-way
interaction terms are also shared, which is not the case in
\tatec. Indeed, such joint parameterization might reduce the
expressiveness of the 2-way interaction terms which, as we
argue in Section \ref{sec:interpretation}, should be left with maximum degrees
of freedom. The \ntn has a more general parameterization than \lfm,
but still uses the same entity embeddings for the 2-way and 3-way
interaction terms. Also, \ntn has two layers and a non-linearity after
the first layer, while our model does not add any nonlinearity after
the embedding step.  In order to have a more precise overview of the
differences between the approaches, we give in Section \ref{TATEC}
(Table \ref{tab:models}) the formulas of the scoring functions of
these related works.



While there has been a lot of focus recently on algorithms purely
based on learning embeddings for entities and/or relationships, many
earlier alternatives had been proposed. We discuss works carried ou
in the Bayesian clustering framework, as well as approaches that
explicitly use the graph structure of the data. The Infinite
Relational Model of \cite{Kemp:2006}, which is a nonparametric
extension of the Stochastic Blockmodel \cite{wang87}, is a Bayesian
clustering approach that learns clusters of entities of the same kind,
i.e. groups of entities that can have similar relationships with other
entities. This work was followed by \cite{Sutskever:2009}, a 3-way
tensor factorization model based on Bayesian clustering in which
entities within a cluster share the same distribution of embeddings.

The last line of work we discuss here are the approaches that use the
structure of the graph in a symbolic way. In that line of work, the
Path Ranking Algorithm ({\sf PRA}) \cite{lao2011random} estimates the
probability of an unobserved fact as a function of the different paths
that go from the subject to the object in the multi-relational graph;
learning consists in finding, for each relationship, a weight
associated to a kind of path (represented as a sequence of
relationships) linking two entities. The {\sf PRA} is used in the
Knowledge Vault project \cite{dong2014knowledge} in conjunction with
an embedding approach. Thus, even though we do not consider these
symbolic approaches here, they could also be combined with our
embedding model if desired.

\section{TATEC}
\label{TATEC}
We now describe our model and the motivations underlying our parameterization.

\subsection{Scoring function}

The data $\calS$ is a set of relations between entities in a fixed set
of entities in $\ents=\{\ent^1, ..., \ent^{\nents}\}$. Relations are
represented as triples $\trp$ where the head $\hd$ and the tail $\tl$
are indexes of entities (i.e. $\hd, \tl \in \intint{\nents}=\{1, ...,
\nents\}$), and the label $\lbl$ is the index of a relationship in
$\labs=\{\lab^1, ..., \lab^\nlabs\}$, which defines the type of the
relation between the entities $\ent^\hd$ and $\ent^\tl$. Our goal is
to learn a discriminant scoring function on the set of all possible
triples $\ents\times\labs\times\ents$ so that the triples which
represent likely relations receive higher scores than triples that
represent unlikely ones. Our proposed model, \tatec, learns embeddings of
entities in a low dimensional vector space, say $\Re^d$, and parameters of operators
on $\Re^d\times\Re^d$, most of these operators being associated to a single
relationship. More precisely, the score given by \tatec to a triple $\trp$, denoted
by $\sco{\trp}$, is defined as:
\begin{equation}
\label{eq:biplustri}
\sco{\trp}=\scorb{\trp}+\scort{\trp}
\end{equation}
where $\scorb{}$ and $\scort{}$ have the following form:
\begin{description}
\item[\citebi] {\sf Bigram} or the $2$-way interaction
  term: $$\scorb{\trp} =
  \dotb{\rhb}{\ehb}+\dotb{\rtb}{\etb}+\dott{\ehb}{\diago}{\etb}\,,$$
  where $\ehb, \etb$ are embeddings in $\Re^\dbi$ of the head and tail
  entities of $\trp$ respectively, $\rhb$ and $\rtb$ are vectors in
  $\Re^\dbi$ that depend on the relationship $\lbl$, and $\diago$
  is a diagonal matrix that does not depend on the input triple.

  As a general notation throughout this section, $\dotb{.}{.}$ is the
  canonical dot product, and $\dott{\bf x}{\bf A}{\bf y} = \dotb{\bf
    x}{\bf Ay}$ where $\bf x$ and $\bf y$ are two vectors in the same
  space and $\bf A$ is a square matrix of appropriate dimensions.

\item[\citetri] {\sf Trigram} or the $3$-way interaction
  term: $$\scort{\trp} = \dott{\eht}{\rt}{\ett}\,,$$ where $\rt$ is a
  matrix of dimensions $(\dtri, \dtri)$, and $\eht$ and $\ett$ are
  embeddings in $\Re^\dtri$ of the head and tail entities
  respectively. The embeddings of the entities for this term are not
  the same as for the $2$-way term; they can even have different dimensions.
\end{description}
The embedding dimensions $\dbi$ and $\dtri$ are hyperparameters of our
model. All other vectors and matrices are learned without any
additional parameter sharing.

The $2$-way interaction term of the model is similar to that of
\cite{bordesMLJ13}, but slightly more general because it does not
contain any constraint between the relation-dependent vectors $\rhb$
and $\rtb$. It can also be seen as a relaxation of the translation
model of \cite{bordesNIPS13}, which is the special case where
$\rhb=-\rtb$, $\diago$ is the identity matrix, and the
entity embeddings are constrained to lie on the unit sphere.

The $3$-way term corresponds exactly to the model used by the
collective factorization method \rescal \cite{Nickel:2011}, and we
chose it for its high expressiveness on complex relationships. Indeed,
as we said earlier in the introduction, $3$-way models can basically
represent any kind of interaction among entities.
The combination of $2$- and $3$-way terms has already
been used in \cite{jenatton2012latent,SocherChenManningNg2013}, but,
besides a different parameterization, \tatec contrasts with them by the
additional freedom brought by using different embeddings in the two interaction terms. In
\lfm \cite{jenatton2012latent}, constraints were imposed on the
relation-dependent matrix of the $3$-way terms (low rank in a limited
basis of rank-one matrices), the relation vectors $\rhb$ and $\rtb$
were constrained to be in the image of the matrix
($\diago=\boldsymbol{0}$ in their work). These global constraints
severely limited the expressiveness of the $3$-way model, and act as
a stringent regularization that reduces the expressiveness of the $2$-way model, which,
as we explain in Section \ref{sec:interpretation}, should
be left with maximum degrees of freedom. We are similar to \ntn \cite{SocherChenManningNg2013}
in the respect that we do not share any parameter between relations. Our overall scoring function is
similar to this model with a single layer, with the fundamental
difference that we use different embedding spaces and do not use any
non-linear transfer function, which results in a facilitated training
(for instance, the gradients have a larger magnitude).

\subsection{Term combination}

We study two strategies for combining the bigram and trigram scores as indicated
in Equation (\ref{eq:biplustri}). In both cases, both $\scorb{}$ and
$\scort{}$ are first trained separately as we detail in
Section~\ref{train} and then combined. The difference between our two
strategies depends on whether we jointly update (or fine-tune) the parameters
of $\scorb{}$ and $\scort{}$ in a second phase or not.

\paragraph{Fine tuning} This first strategy, denoted
\tatecft, simply consists in summing both scores following 
Equation (\ref{eq:biplustri}). 
\begin{equation*}
\sco_{FT}{\trp}= \dotb{\rhb}{\ehb}+\dotb{\rtb}{\etb}+\dott{\ehb}{\diago}{\etb}+\dott{\eht}{\rt}{\ett}
\end{equation*}
All parameters of $\scorb{}$ and $\scort{}$ (and hence of $\sco{}$)
are then fine-tuned in a second training phase to accommodate for their
combination. This version could be trained directly without
pre-training $\scorb{}$ and $\scort{}$ separately but we show in our
experiments that this is detrimental.

\paragraph{Linear combination} The second strategy combines the
bigram and trigram terms using a linear combination, without jointly
fine-tuning their parameters that remain unchanged after their pre-training. The score $\sco{}$  is hence defined as follows:
\begin{equation*}
\sco_{LC}{\trp} =\delta_1^\ell \dotb{\rhb}{\ehb}+\delta_2^\ell \dotb{\rtb}{\etb}+\delta_3^\ell \dott{\ehb}{\diago}{\etb} + \delta_4^\ell \dott{\eht}{\rt}{\ett}
\end{equation*}
The combination weights $\delta_i^\ell$ depend on the relationship and
are learned by optimizing the ranking loss (defined later in
\eqref{eq:loss}) using L-BFGS, with an additional quadratic
penalization term, $\sum_{\ell}
\frac{||\boldsymbol{\delta}^\ell||_2^2}{\sigma_\ell + \epsilon}$,
where $\boldsymbol{\delta}^\ell$ contains the combination weights for
relation $\ell$, and $\sigma$ are constrained to $\sum_{\ell}
\sigma_\ell = \alpha$ ($\alpha$ is a hyperparameter).
%
%
This version of \tatec is denoted \tateclc in the following.

\subsection{Interpretation and motivation of the model}
\vspace*{-0.5ex}
\label{sec:interpretation}
This section discusses the motivations underlying the parameterization
of \tatec, and in particular our choice of  $2$-way model to
complement the $3$-way term.

\subsubsection{$2$-way interactions as one fiber biases} 

It is common in regression, classification or collaborative filtering
to add biases (also called offsets or intercepts) to the model. For
instance, a critical step of the best-performing techniques of the Netflix
prize was to add user and item biases, i.e. to approximate a
user-rating $R_{ui}$ according to (see e.g. \cite{Koren09}):
\begin{equation}
\label{eq:collabfilt}
R_{ui} \approx \dotb{{\bf P}_u}{ {\bf Q}_i} + \alpha_u+\beta_i+\mu
\end{equation}
where ${\bf P}\in\Re^{U\times k}$, with each row ${\bf P}_u$
containing the $k$-dimensional embedding of the user ($U$ is the
number of users), ${\bf Q}\in\Re^{I\times k}$ containing the
embeddings of the $I$ items, $\alpha_u\in\Re$ a bias only depending on
a user and $\beta_i\in\Re$ a bias only depending on an item ($\mu$ is a
constant that we do not consider further on).

The $2$-way + $3$-way interaction model we propose can be seen as the
3-mode tensor version of this ``biased'' version of matrix
factorization: the trigram term \citetri is the collective matrix
factorization parameterization of the \rescal algorithm
\cite{Nickel:2011} and plays a role analogous to the term $\dotb{{\bf
    P}_u}{ {\bf Q}_i}$ of the matrix factorization model for
collaborative filtering \eqref{eq:collabfilt}. The bigram term
\citebi then plays the role of biases for each fiber of the
tensor,\footnote{Fibers are the higher order analogue of matrix rows
  and columns for tensors and are defined by fixing every index but one.}
i.e.
\begin{equation}
\label{eq:biasbiases} 
\scorb{\trp} \approx B^1_{\lab,\hd}+B^2_{\lab,\tl}+B^3_{\hd, \tl}
\end{equation}
and thus is the  analogue for tensors to the term $\alpha_u+\beta_i$ in the
matrix factorization model \eqref{eq:collabfilt}. The exact form of
$\scorb{\trp}$ given in \citebi corresponds to a specific form of
collective factorization of the fiber-wise bias matrices ${\bf
  B}^1=\left[B^1_{\lab,\hd}\right]_{\lab\in\intint{\nlabs},
  \hd\in\intint{\nents}}$, ${\bf B}^2$ and ${\bf B}^3$ of Equation
(\ref{eq:biasbiases}).
We do not exactly learn one bias by fiber because many such fibers
have very little data, while, as we argue in the following, the
specific form of collective factorization we propose in \citebi
should allow to share relevant information between different biases.

\subsubsection{The need for multiple embeddings} 
A key feature of \tatec is to use different embedding spaces for the
$2$-way and $3$-way terms, while existing approaches that have both
types of interactions use the same embedding space
\cite{jenatton2012latent,SocherChenManningNg2013}. We motivate this
choice in this section.

It is important to notice that biases in the matrix factorization
model \eqref{eq:collabfilt}, or the bigram term in the overall scoring
function \eqref{eq:biplustri} do not affect the model expressiveness,
and in particular do not affect the main modeling assumptions that
embeddings should have low rank. The user/item-biases in
\eqref{eq:collabfilt} only boil down to adding two rank-$1$ matrices
$\boldsymbol{\alpha}\boldsymbol{1}^T$ and
$\boldsymbol{1}\boldsymbol{\beta}^T$ to the factorization model. Since
the rank of the matrix is a hyperparameter, one may simply add $2$ to
this hyperparameter and get a slightly larger expressiveness than
before, with reasonably little impact since the increase in rank would
remain small compared to its original value (which is usually $50$ or $100$
for large collaborative filtering data sets). The critical feature of
these biases in collaborative filtering is how they interfere with
capacity control terms other than the rank, namely the $2$-norm
regularization: in \cite{Koren09} for instance, all terms of
\eqref{eq:collabfilt} are trained using a squared error as a measure of
approximation and regularized by $\lambda\left(\norm{{\bf
      P}_u}^2+\norm{{\bf Q}_i}^2+\alpha_u^2+\beta_i^2\right)$, where
$\lambda>0$ is the regularization factor. This kind of regularization
is a weighted trace norm regularization \cite{salakhutdinov10} on
${\bf P}{\bf Q}^T$. Leaving aside the ``weighted" part, the idea is
that at convergence, the quantity $\lambda\left(\sum_u\norm{{\bf
      P}_u}^2+\sum_i\norm{{\bf Q}_i}^2\right)$ is equal to $2\lambda$
times the sum of the singular values of the matrix ${\bf P}{\bf
  Q}^T$. However, $\lambda \norm{\boldsymbol{\alpha}}^2$, which is the
regularization applied to user biases, is \emph{not} $2\lambda$ times
the singular value of the rank-one matrix
$\boldsymbol{\alpha}\boldsymbol{1}^T$, which is equal to
$\sqrt{I}\norm{\boldsymbol{\alpha}}$, and can be much larger than
$\norm{\boldsymbol{\alpha}}^2$. Thus, if the pattern user+item biases
exists in the data, but very weakly because it is hidden by stronger
factors, it will be less regularized than others and the
model should be able to capture it. 
Biases,  which are allowed to fit the data
more than other factors, 
offer the opportunity of relaxing the control of capacity on some
parts of the model but this translates into gains if the patterns that
they capture are indeed useful patterns for
generalization. Otherwise, this ends up relaxing the capacity to lead
to more overfitting.

Our bigram terms are closely related to the trigram term: the terms
$\dotb{\rhb}{\ehb}$ and $\dotb{\rtb}{\etb}$ can be added to the
trigram term by adding constant features in the entities' embeddings,
and $\dott{\ehb}{\diago}{\etb}$ is directly in an appropriate
quadratic form. Thus, the only way to gain from the addition of bigram
terms is to ensure that they can capture useful patterns, but also
that capacity control on these terms is less strict than on the
trigram terms. 
In tensor factorization models, and especially $3$-way
interaction models with parameterizations such as \citetri, capacity
control through the regularization of individual parameters is still
not well understood, and sometimes turns out to be more detrimental than effective in experiments. The only effective parameter is the
admissible rank of the embeddings, which leads to the conclusion that
the bigram term can be really useful in addition to the trigram term
if higher-dimensional embeddings are used. Hence, in absence of clear and concrete way of effectively
controlling the capacity of the trigram term, we believe that different embedding
spaces should be used.

\subsubsection{$2$-way interactions as entity types+similarity}

Having a part of the model that is less expressive, but less
regularized (see Subsection \ref{regularization}) than the other part is only useful if the patterns it
can learn are meaningful for the prediction task at hand. In this
section, we give the motivation for our $2$-way interaction term for
the task of modeling multi-relational data.

Most relationships in multi-relational data, and in knowledge bases
like \fb in particular, are strongly typed, in the sense that only
well-defined and specific subsets of entities can be either heads or
tails of selected relationships. For instance, a relationship like
{\tt capital\_of} expects a (big) city as head and a country as tail
for any valid relation. Large knowledge bases have huge amounts of
entities, but those belong to many different types. Identifying the
expected types of head and tail entities of relationships, with an
appropriate granularity of types (e.g.  {\tt person} or {\tt artist} or
{\tt writer}), is likely to filter out $95\%$ of the entity set during
prediction. 
The exact form of the first two terms
$\dotb{\rhb}{\ehb}+\dotb{\rtb}{\etb}$ of the $2$-way interaction model
\citebi, which corresponds to a low-rank factorization
of the per bias matrices (\su, \ve) and (\ob, \ve) in
which \su and \ob entities have the same embeddings, is based on the assumption that the types of entities can be predicted based on few (learned) features,
and these features are the same for predicting \su-types as for predicting \ob-types. As such, it is natural to share the entities embeddings in the first two
terms of \citebi.

The last term, $\dott{\ehb}{\diago}{\etb}$, is intended to account for
a global similarity between entities. For instance, the capital of France can easily be predicted
by looking for the city with strongest overall connections with France in 
the knowledge base. A country and a city may be strongly linked
through their geographical positions, independent of their respective
types. The diagonal matrix $\diago$ allows to re-weight features of the
embedding space to account for the fact that the features used to
describe types may not be the same as those that can describe the
similarity between objects of different types. The use of a diagonal
matrix is strictly equivalent to using a general symmetric matrix in
place of $\diago$.\footnote{We can see the equivalence by taking the
  eigenvalue decomposition of a symmetric $\diago$: apply the change of
  basis to the embeddings to keep only the diagonal part of $\diago$ in
  the term $\dott{\ehb}{\diago}{\etb}$, and apply the reverse
  transformation to the vectors $\rhb$ and $\rtb$. Note that since
  rotations preserve Euclidean distances, the equivalence still holds
  under $2$-norm regularization of the embeddings.} The reason for
using a symmetric matrix comes from the intuition that the direction
of many relationships is arbitrary (i.e. the choice between having
triples ``Paris is capital of France" rather than ``France has capital
Paris"), and the model should be invariant under arbitrary inversions
of the directions of the relationships (in the case of an inversion of
direction, the relations vectors $\rhb$ and $\rtb$ are swapped, but
all other parameters are unaffected). For tasks in which such
invariance is not desirable, the diagonal matrix could be replaced by an
arbitrary matrix.

\begin{table}[t]
  \caption{\label{tab:models}{\bf Scoring function for several models
      of the literature.} Capitalized letters denote matrices and
    lower cased ones, vectors.}
\begin{center}
\vspace*{-1ex}
\begin{small}
\begin{sc}
\begin{tabular}{|c|c |}
\hline
Model & Score ($\sco\trp$) \\
\hline
\transe & $||\eh+\rel-\et||_2$ \\
\transh & $||(\eh-\dotb{\w}{\eh \w})+\rel-(\et-\dotb{\w}{\et \w})||_2^2$ \\
\transr & $||\dotb{\eh}{\bf{M}_\lbl}+\rel-\dotb{\et}{\bf{M}_\lbl}||_2$\\
\rescal & $\dott{\eh}{\rt}{\et}$\\
\lfm & $\dott{y}{\rt}{y'}+\dott{\eh}{\rt}{{\bf z}}+\dott{{\bf z}}{\rt}{\et} + \dott{\eh}{\rt}{\et}$\\
\hline
\end{tabular}
\end{sc}
\end{small}
\end{center}
\vspace*{-4ex}
\end{table}

\section{Training}
\label{train}

\subsection{Ranking objective}
\label{ranking}
Training \tatec is carried out using stochastic gradient descent over
a ranking objective function, which is designed to give higher scores
to positive triples (facts that express true and verified information
from the KB) than to negative ones (facts that are supposed to express false information). These negative triples can be provided by the KB, but often they are not, so we need a process to turn positive triples into corrupted ones to carry out our discriminative training. A simple approach consists in creating negative examples by replacing one argument of a positive triple by a random element. This way is simple and efficient in practice but may introduce noise by creating wrong negatives.

Let $\calS$ be the set of positive triples provided by the KB, we optimize the following ranking loss function:
\begin{equation}
\label{eq:loss}
\sum_{\trp\in\calS}\sum_{\trpprime\in{\cal C}\trp } \big [\gamma-\sco{\trp}+\sco{\trpprime}\big]_+
\end{equation}
where $[z]_+ = \max(z, 0)$ and ${\cal C}(\hd,\lbl, \tl$) is the set of corrupted triples. Depending on the application, this set can be defined in 3 different ways:
\begin{enumerate}
\item ${\cal C}\trp = \{\trpprime \in \intint{\nents}\times\labs\times\intint{\nents} | \hdprime\neq\hd \text{~and~} \lblprime\neq\lbl\ \text{~and~} \tlprime\neq\tl \}$
\item ${\cal C}\trp = \{\trpprimee \in \intint{\nents}\times\labs\times\intint{\nents} | \hdprime\neq\hd \text{~or~} \tlprime\neq\tl\}\,$
\item ${\cal C}\trp = \{\trpprimeee \in \intint{\nents}\times\labs\times\intint{\nents} | \lblprime\neq\lbl\}\,$
\end{enumerate}

The margin $\gamma$ is an hyperparameter that defines the minimum gap between the score of a positive triple and its negative one's. The stochastic gradient descent is performed in a minibatch setting. At each epoch the data set is shuffled and split into disjoint minibatches of $m$ triples and $1$ or $2$ (see next section) negative triples are created for every positive one. We use two different learning rates $\lambda_1$ and $\lambda_2$, one for the \bi and one the \tri model; they are kept fixed during the whole training.

We are interested in both \bi and \tri terms of \tatec to capture
different data patterns, and using a random initialization of all
weights can not necessarily lead to such a solution. Hence, we first
pre-train separately $\scorb{\trp}$ and $\scort{\trp}$, and then we
use these learned weights to initialize that of the full
model. Training of \tatec is hence carried out in two phases: a
(disjoint) pre-training and either a (joint) fine-tuning for \tatecft
or a learning of the combination weights for \tateclc. Both
pre-training and fine-tuning are stopped using early stopping on a
validation set, and follow the training procedure that is
summarized in Algorithm \ref{alg:tatec}, for the unregularized case.
Training of the linear combination weights of \tateclc is stopped at convergence of L-BFGS.

\begin{algorithm}[t]
\caption{Learning unregularized \tatec. }
\begin{small}
\label{alg:tatec}
  \begin{algorithmic}[1]
     	\INPUT Training set $S=\{(h,l,t)\}$, margin $\gamma$, learning rates $\lambda_1$ and $\lambda_2$
        \STATE {\bf initialization} 
        \STATE ~~~~~-~for \bi: ${\bf e}_1 \leftarrow$ uniform$(-\frac{6}{\sqrt{d_1}}, \frac{6}{\sqrt{d_1}})$ for each entity $e$ 
	\STATE ~~~~~~~~~~~~~~~~~~~~~~~~~ ${\bf r}_1, {\bf r}_2 \leftarrow $ uniform$(-\frac{6}{\sqrt{d_1}}, \frac{6}{\sqrt{d_1}})$ for each $\ell$ 
	\STATE ~~~~~~~~~~~~~~~~~~~~~~~~~ ${\bf D} \leftarrow $ uniform$(-\frac{6}{\sqrt{d_1}}, \frac{6}{\sqrt{d_1}})$ 
        \STATE ~~~~~-~for \tri: ${\bf e}_2 \leftarrow$ uniform$(-\frac{6}{\sqrt{d_2}}, \frac{6}{\sqrt{d_2}})$ for each entity $e$ 
	\STATE ~~~~~~~~~~~~~~~~~~~~~~~~~ ${\bf R} \leftarrow $ uniform$(-\frac{6}{\sqrt{d_2}}, \frac{6}{\sqrt{d_2}})$ for each $\ell$ 
	\STATE ~~~~~-~for \tatecft: pre-trained weights of \bi and \tri
	\STATE All the embeddings are normalized to have a 2- or Frobenius-norm equal to 1.
	\LOOP
        \STATE $S_{batch} \leftarrow sample(S, m)$ // sample a training
        minibatch of size $m$
        \STATE $T_{batch} \leftarrow \emptyset $ // initialize a set of
        pairs of examples
	\FOR{$(h,\ell,t) \in S_{batch}$}
                \STATE $(h',\ell',t') \leftarrow $ sample a negative
                triple according to the selected strategy ${\cal C}\trp$
	     	\STATE $ T_{batch} \leftarrow
                T_{batch}\cup\left\{\big((h,\ell,t),(h',\ell',t')
                  \big)\right\}$ // record the pairs of examples

	\ENDFOR
	\STATE Update parameters using gradients $\displaystyle\sum_{\big((h,\ell,t),(h',\ell',t')\big)
          \in T_{batch}} \hspace{-0.6cm}\nabla
        \big[\gamma-\sco{\trp}+\sco{\trpprime}\big]_+$:
	\STATE ~~~~~-~for \bi: $\sco = \sco_1$
	\STATE ~~~~~-~for \tri: $\sco = \sco_2$
	\STATE ~~~~~-~for \tatecft: $\sco = \sco_1+\sco_2$
	\ENDLOOP
   \end{algorithmic}
\end{small}
\end{algorithm}

\subsection{Regularization}
\label{regularization}
Previous work on embedding models have used two different
regularization strategies: either by constraining the entity
embeddings to have, at most, a $2$-norm of value $\rho_e$
\cite{duranECML} or by adding a $2$-norm penalty on the weights
\cite{wang2014knowledge,lin2015learning} to the objective function
\eqref{eq:loss}. In the former, which we denote as {\it hard
  regularization}, regularization is performed by projecting the
entity embeddings after each minibatch onto the 2-norm ball of radius
$\rho_e$. In the latter, which we denote as {\it soft regularization},
a penalization term of the form $[||{\bf e}||_2^2 - \rho_e^2]_+$ for the
entity embeddings ${\bf e}$ is added. The soft scheme allows the $2$-norm
of the embeddings to grow further than $\rho_e$, with a penalty.

To control the large capacity of the relation matrices in the \tri
model, we have adapted the two regularization schemes: in the {\it
  hard} scheme, we force the relation matrices to have, at most, a
Frobenius norm of value $\rho_l$, and in the {\it soft} one, we
include a penalization term of the form $[||{\bf R}||_F^2 - \rho_l^2]_+$ to
the loss function (\ref{eq:loss}) .
As a result, in the {\it soft} scheme the following regularization term is added to
the loss function (\ref{eq:loss}):
$C_1 [||{\bf e}_1||_2^2 - \rho_e^2]_+ + C_2 \big([||{\bf e}_2||_2^2 - \rho_e^2]_+ +
[||{\bf R}||_F^2 - \rho_l^2]_+\big)$,
where $C_1$ and $C_2$ are hyperparameters that weight the importance
of each soft constraint. In terms of practicality, the bigger
flexibility of the soft version comes with one more
hyperparameter.
In the following, the suffixes {\scriptsize soft} and {\scriptsize
  hard} are used to refer to either of those regularization scheme.
\tatec has also an other implicit regularization factor since it is using the same entity representation for an entity
regardless of its role as head or tail.

To sum up, in the hard regularization case, the optimization problem
for \tatecft is:
\begin{equation*}
\begin{aligned}
\label{eq:lossHARD}
\min &\sum_{\trp\in\calS}\sum_{\trpprime\in\cortrp} \big[\gamma-\sco{\trp}+\sco{\trpprime}\big]_+\\
&\text{s.t.}\quad ||{\bf e}^i_1||_2 \leq \rho_e \quad\forall i \in \intint{\nents} \\
& ~~\quad\quad ||{\bf e}^ i_2||_2 \leq \rho_e \quad\forall i \in \intint{\nents} \\
& ~~\quad\quad ||R^\ell||_F \leq \rho_l \quad\forall \ell \in \intint{\nlabs} \\
\end{aligned}
\end{equation*}

And in the soft regularization case it is:
\begin{equation*}
\begin{aligned}
\label{eq:lossSOFT}
\min &\sum_{\trp\in\calS}\sum_{\trpprime\in\cortrp} [\gamma-\sco{\trp}+\sco{\trpprime}]_+ + C_1\sum_{i \in \intint{\nents}} [||{\bf e}_1^i||_2^2 - \rho_e^2]_+\\
& + C_2\big(\sum_{i \in \intint{\nents}} [||{\bf e}_2^i||_2^2 - \rho_e^2]_+ + \sum_{\ell \in \intint{\nlabs}} [||{\bf R}^\ell||_F^2 - \rho_l^2]_+\big)
\end{aligned}
\end{equation*}
where $\sco{\trp}=\dotb{\rhb}{\ehb}+\dotb{\rtb}{\etb}+\dott{\ehb}{\diago}{\etb}+\dott{\eht}{\rt}{\ett}$ in both cases.

\section{Experiments}
\label{results}
This section presents various experiments to illustrate how competitive \tatec is with respect to several state-of-the-art models on 4 benchmarks from the literature: \umls, \kinships, \fb and \svo. The statistics of these data sets are given in Table \ref{tab:StatsKB}. 
All versions of \tatec and of its components \bi and \tri are compared with the state-of-the-art models for each database.


\begin{table}[t]
\caption{{\bf Statistics of the data sets} used in this paper and extracted from four knowledge bases: \fb, \svo, \kinships and \umls.}
\label{tab:StatsKB}
\begin{small}
\begin{center}
\begin{sc}
\vspace*{-1ex}
\begin{tabular}{|l|r|r|r|r|}
\hline
Data set & \fb & \svo & \kinships & \umls \\
\hline
Entities    & 14,951 & 30,605 & 104 & 135\\
Relationships & 1,345 & 4,547 & 26 & 49\\
Training examples    & 483,142 & 1,000,000 & 224973 & 102612\\
Validation examples    & 50,000 & 50,000 & 28122 & 89302 \\
Test examples     & 59,071 & 250,000 & 28121 & 89302\\
\hline
\end{tabular}
\vspace*{-5ex}
\end{sc}
\end{center}
\end{small}
\end{table}

\subsection{Experimental setting}

This section details the protocols used in our various experiments.

\subsubsection{Datasets and metrics}

Our experimental settings and evaluation metrics are borrowed from previous works, so as to allow for result comparisons.

\paragraph{\umls/\kinships}
\kinships \cite{denham73} is a KB expressing the relational structure of the kinship system of the Australian tribe Alyawarra, and \umls \cite{mccray2003} is a KB of biomedical high-level concepts like diseases or symptoms connected by verbs like \texttt{complicates}, \texttt{affects} or \texttt{causes}. 
For these data sets, the whole set of possible triples, positive or
negative, is observed. We used the area under the precision-recall
curve as metric. The dataset was split in 10-folds for cross-validation: 8 for training, 1 for validation and the last one for test. Since the number of available negative triples is much bigger than the number of positive triples, the positive ones of each fold are replicated to match the number of negative ones. These negative triples correspond to the first setting of negative examples of Section \ref{ranking}. The number of training epochs was fixed to 100. \bi, \tri and \tatec models were validated every 10 epochs using the AUC under the precision-recall curve as validation criterion over 1,000 randomly chosen validation triples - keeping the same proportion of negative and positive triples. For \transe, which we ran as baseline, we validated every 10 epochs as well.

\paragraph{\fb}
Introduced in \cite{bordesNIPS13}, this data set is a subset of {\sc
  Freebase}, a very large database of generic facts gathering more than 1.2 billion triples and 80 million entities.
For evaluation on it, we used a ranking metric. The head of each test triple is replaced by each of the entities of the dictionary in turn, and the score is computed for each of them. These scores are sorted in descending order and the rank of the correct entity is stored. The same procedure is repeated when removing the tail instead of the head. The mean of these ranks is the \textit{mean rank}, and the proportion of correct entities ranked in the top 10 is the \textit{hits@10}. This is called the {\it raw} setting. In this setting correct positive triples can be ranked higher than the target one and hence be counted as errors. Following \cite{duranECML}, in order to reduce this noise in the measure, and thus granting a clearer view on ranking performance, we remove all the positive triples that can be found in either the training, validation or testing set, except the target one, from the ranking. This setting is called {\it filtered}.

Since \fb is made up only of positive triples, the negative ones have to be generated. To do that, in each epoch we generate two negative triples per positive by replacing a single unit of the positive triple by a random entity (once the head and once the tail). This corruption approach implements the prior knowledge that unobserved triples are likely to be invalid, and has been widely used in previous work when learning embeddings of knowledge bases or words in the context of language models. These negative triples correspond to the second setting of negative examples of Section \ref{ranking}.
We ran 500 training epochs for both \transe, \bi, \tri and \tatec, and using the final filtered mean rank as validation criterion. If several models statistically have similar filtered mean ranks, we take the $hits@10$ as secondary validation criterion.\footnote{Results on both \fb and \svo with \transe and \tatec are provided in \cite{duranECML}, however in these works the hyperparameters were validated on a smaller validation set, that led to suboptimal results.}
Since for this dataset, training, validation and test sets are fixed, to give a confidence interval to our results, we randomly split the test set into 4 subsets before computing the evaluation metrics. We do this 5 times, and finally we compute the mean and the standard deviation over these 20 values for mean rank and hits@10.

\paragraph{\svo}
\svo is a database of nouns connected by verbs through subject-verb-direct object relations and extracted from Wikipedia articles. It has been introduced in \cite{jenatton2012latent}.
For this database we perform a verb prediction task, where one has to assign the correct verb given two nouns acting as subject and direct object; in other words, we present results of ranking \textit{label} given \textit{head} and \textit{tail}. As for \fb, two ranking metrics are computed, the \textit{mean rank} and the \textit{hits@5\%}, which is the proportion of predictions for which the correct verb is ranked in the top 5$\%$ of the total number of verbs, that is within the top 5$\%$ of 4,547 $\approx$ 227. We use the {\it raw} setting for \svo.
Due to the different kind of task (predicting  \textit{label} instead of predicting  \textit{head}/\textit{tail}), the negative triples have been generated by replacing the label by a random verb. These negative triples correspond to the third setting of negative examples of Section \ref{ranking}. For \transe, \bi and \tri the number of epochs has been fixed to 500 and they were validated every 10 epochs.  For \tatec we ran only 10 epochs, and validated for each. The mean rank has been chosen as validation criterion over 1,000 random validation triples.


\subsubsection{Implementation}
To pre-train our \bi and \tri models we validated the learning rate
for the stochastic gradient descent among
$\{0.1, 0.01, 0.001, 0.0001\}$ and the margin among $\{0.1, 0.25,$
$0.5, 1\}$. The radius $\rho_e$ determining the value from which the
$L_2$-norm of the entity embeddings are penalized has been fixed to 1,
but the radius $\rho_l$ of the \tri model has been validated among
$\{0, 1, 5, 10, 20\}$. Due to the different size of these KBs, the
embedding dimension $d$ has been validated in different ranges. For
\svo it has been selected among $\{25, 50\}$, among $\{50, 75, 100\}$
for \fb and among $\{10, 20, 40\}$ for \umls and \kinships. When the
soft regularization is applied, the regularization parameter has been validated
among $\{0, 0.0001, 0.001, 0.01, 0.1, 1, 10, 100\}$. For fine-tuning
\tatec, the learning rates were selected among the same values for
learning the \bi and \tri models in isolation, independent of the
values chosen for pre-training, and so are the margin and for the
penalization terms $C_1$ and $C_2$ if the soft regularization is
used. The configurations of the model selected using their performance
on the validation set are given in Appendix~\ref{app:hp}.

Training of the combination weights of \tateclc is carried out in an iterative way, by alternating optimization of $\delta$ parameters via L-BFGS, and update of $\sigma$ parameters using $\sigma_\ell^* = \frac{\alpha ||\boldsymbol{\delta}^\ell||_2}{\sum_{k}||\boldsymbol{\delta}^k||_2}$, until some stopping criterion is reached. The $\delta$ parameters are initialized to $1$ and the $\alpha$ value is validated among $\{0.1, 1, 10, 50, 100,
200, 500, 1000\}$. 

\subsubsection{Baselines}

\paragraph{Variants}

We performed breakdown experiments with 2 different versions of \tatec to assess the impact of its various aspects. These variants are:
\begin{itemize}
\item {\sc \tatecft-no-pretrain}: \tatecft without pre-training $s_1(h,l,t)$ and $s_2(h,l,t)$.
\item {\sc \tatecft-shared}: \tatecft but sharing the entities embeddings between $s_1(h,l,t)$ and $s_2(h,l,t)$ and without pre-training.
\end{itemize}

The experiments with these 3 versions of \tatec have been performed in the soft regularization setting. Their hyperparameters were chosen using the same grid as above.

\paragraph{Previous models}

We retrained \transe ourselves with the same hyperparameter grid as for \tatec and used it as a running baseline on all datasets, using either soft or hard regularization. In addition, we display the results of the best performing methods of the literature on each dataset, with values extracted from the original papers.

On \umls and \kinships, we also report the performance of the 3-way models \rescal, \lfm and the 2-way \smel.
On \fb, recent variants of \transe, such as \transh, \transr and \ctransr \cite{lin2015learning} have been chosen as main baselines.  Both in \transh and \transr/\ctransr, the optimal values of the hyperparameters as the dimension, the margin or the learning rate have been selected within similar ranges as those for \tatec. 
%
%
On \svo, we compare \tatec with three different approaches: {\sc
  Counts}, the 2-way model \smel and the 3-way \lfm. {\sc Counts} is
based on the direct estimation of probabilities of triples (\su, \ve,
\ob) by using the number of occurrences of pairs (\su, \ve) and (\ve,
\ob) in the training set. The results for these models have been
extracted from \cite{jenatton2012latent}, and we followed their
experimental setting. Since the results in this paper are only
available in the raw setting, we restricted our experiments to this
configuration on \svo as well.

\begin{table}[t]
 \caption{{\bf Test AUC under the precision-recall curve on \umls and \kinships} {\small for models from the literature (top) and 
\tatec (bottom).. Best performing methods are in bold.}}
\label{tab:resultsUMKI}
\begin{center}
\vspace*{-2ex}
\begin{small}
\begin{sc}
\begin{tabular}{|l|r@{\:}l||r@{\:}l|}
\hline
Model &\multicolumn{2}{c|}{\umls} & \multicolumn{2}{c|}{\kinships} \\
\hline
\hline 
\smel & 0.983 & $\pm$ 0.003& 0.907 & $\pm$ 0.008\\
\rescal & 0.98 & & {\bf 0.95} &  \\
\lfm & {\bf 0.990} & $\pm$ 0.003& {\bf 0.946} & $\pm$ 0.005\\
\transe\soft & 0.734 & $\pm$ 0.033 & 0.135 & $\pm$ 0.005\\
\transe\hard & 0.706 & $\pm$ 0.034 & 0.134 & $\pm$ 0.005\\
\hline
\hline
\bi\hard& 0.936 & $\pm$ 0.020 & 0.140 & $\pm$ 0.004\\
\tri\hard & 0.980 & $\pm$ 0.006 & {\bf 0.943} & $\pm$ 0.009\\
\hdashline
\tatecft\hard & { 0.984} & $\pm$ 0.004 & 0.876 & $\pm$ 0.012\\
\hline
\bi\soft & 0.936 & $\pm$ 0.018 & 0.141 & $\pm$ 0.003\\
\tri\soft & {0.983} & $\pm$ 0.004 & {\bf 0.948} & $\pm$ 0.008\\
\hdashline
\tatecft\soft & {\bf 0.985} & $\pm$ 0.004 & 0.919 & $\pm$ 0.008\\
\tateclc\soft & {\bf 0.985} & $\pm$ 0.004 & {\bf 0.941} & $\pm$ 0.009\\
\hline 
\end{tabular}
\end{sc}
\end{small}
\end{center}
\end{table}

\begin{table}[t]
 \caption{{\bf Test results on \fb and \svo} {\small for models from the literature (top), 
\tatec (middle) and variants (bottom). Best performing methods are in
bold. The {\it filtered} setting is used for \fb and the {\it raw}
setting for \svo.}}
\label{tab:resultsFBSV}
\begin{center}
\vspace*{-2ex}
\begin{small}
\begin{sc}
\begin{tabular}{|l|r@{\:}l|r@{\:}l|r@{\:}l|r@{\:}l|}
\hline
& \multicolumn{4}{c|}{\fb} & \multicolumn{4}{c|}{\svo} \\
\cline{2-9}
  Model &  \multicolumn{2}{c|}{Mean Rank} &  \multicolumn{2}{c|}{Hits@10} & \multicolumn{2}{c|}{Mean Rank}  & \multicolumn{2}{c|}{Hits@5$\%$} \\
\hline
\hline
\sc Counts & - & & -&  &517.4 & & 72 & \\
\smel & - & & - & &199.6 & &  77 & \\
\lfm & - & & -&  &195 & & 78 & \\
\transh & 87   & & 64.4 &  &- & &- & \\
\transr & 77   & & 68.7 &  &- & &- & \\
\ctransr & 75  & & 70.2 &  &- & &- & \\
\transe\soft & {\bf 50.7} & $\pm$  2.0 & 71.5 & $\pm$ 0.3  & 282.5  & $\pm$ 1.7 &70.6 & $\pm$ 0.2\\
\transe\hard & {\bf 50.6} & $\pm$ 2.0 & 71.5 & $\pm$ 0.3 & 282.8  & $\pm$ 2.3 & 70.6 & $\pm$ 0.2\\
\hline
\hline 
{\sc \tatec-no-pretrain} & 97.1 & $\pm$ 3.9&  65.7 & $\pm$ 0.2 &- & & - & \\
{\sc \tatec-shared} & 94.8 & $\pm$ 3.2 & 63.4 & $\pm$ 0.3  &- & & - &
  \\
\hline 
\bi\hard & 94.5 & $\pm$ 2.9 & 67.5 &  $\pm$ 0.4 & 219.2 &$\pm$ 1.9 & 77.6 & $\pm$ 0.1\\
\tri\hard & 137.7 & $\pm$ 7.1 & 56.1 & $\pm$ 0.4  & 187.9 &$\pm$ 1.2 &
                                                                       79.5 & $\pm$ 0.1\\
\hdashline
\tatecft\hard & 59.8 & $\pm$ 2.6 & {\bf 77.3} &  $\pm$ 0.3 & 188.5 & $\pm$ 1.9 & 79.8 & $\pm$ 0.1\\
\hline 
\bi\soft & 87.7 & $\pm$  4.1 & 70.0 &  $\pm$ 0.2  & 211.9 & $\pm$ 1.8 & 77.8 & $\pm$ 0.1\\
\tri\soft & 121.0 & $\pm$ 7.2 & 58.0 & $\pm$ 0.3  &189.2 & $\pm$ 2.1 & 79.5 & $\pm$ 0.2\\
\hdashline
\tatecft\soft & 57.8 & $\pm$ 2.3 & {\bf 76.7} & $\pm$ 0.3  & 185.4 &
                                                                     $\pm$ 1.5 & {\bf 80.0} &  $\pm$ 0.1\\
\tateclc\soft & 68.5 & $\pm$ 3.2 &  72.8 & $\pm$ 0.2 & {\bf 182.6} &
                                                                     $\pm$ 1.2 & {\bf 80.1} & $\pm$ 0.1  \\
\hline
\end{tabular}
\end{sc}
\end{small}
\end{center}
\end{table}

\subsection{Results}
We recall that the suffixes {\scriptsize soft} or {\scriptsize hard}
refer to the regularization scheme used, and the suffixes {\sc ft} and {\sc
lc} to the combination strategy of \tatec.

\subsubsection{\umls and \kinships}
The results for these two knowledge bases are provided in Table~\ref{tab:resultsUMKI}. 
In \umls, most models are performing well. The combination of the \bi
and \tri models is slightly better than the \tri alone but it is not
significant. It seems that the constituents of \tatec, \bi and \tri,
do not encode very complementary information and their combination
does not bring much improvement. Basically, on this dataset, many
methods are somewhat as efficient as the best one, \lfm. The difference between \transe and \bi on this dataset illustrates the potential impact of the diagonal matrix $\diago$, which does not constrain embeddings of both head and tail entities of a triple to be similar.

Regarding \kinships, there is a big gap between 2-way models like
\transe and 3-way models like \rescal. The cause of this deterioration
comes from a peculiarity of the positive triples of this KB: each
entity appears 104 times -- the number of entities in this KB -- as head
and it is connected to the 104 entities -- even itself -- only once. In other words, the conditional probabilities $P(head|tail)$ and $P(tail|head)$ are totally uninformative. This has a very important consequence for the 2-way models since they highly rely on such information: for \kinships,  the interaction head-tail is, at best, irrelevant, though in practice this interaction may even introduce noise.

Due to the poor performance of the \bi model, when it is combined with
the \tri model this combination can turn out to be detrimental
w.r.t. to the performance of \tri in isolation: 2-way models are quite
noisy for this KB and we cannot take advantage of them. On the other
side the \tri model logically reaches a very similar performance to
\rescal, and similar to \lfm as well.
Performance of \tatec versions based on fine-tuning of the parameters
(\tatecft) are
worse than that of \tri because \bi degrades the model.
\tateclc, using a -- potentially sparse -- linear combination of the
models, does not have this drawback since it can completely cancel out
the influence of bigram model.
 As a conclusion from the experiments in this KB, when one of the
 components of \tatec is quite noisy, we should directly remove it and
 \tateclc can do it automatically. The soft regularization setting
 seems to be slightly better also.

\subsubsection{\fb}
Table \ref{tab:resultsFBSV} (left) displays results on \fb. Unlike for
\kinships, here the 2-way models outperform the 3-way models in both
mean rank and hits@10. The simplicity of the 2-way models seems to be
an advantage in \fb: this is something that was already observed
in cite \cite{yang2014learning}. The combination of the \bi and \tri
models into \tatec leads to an impressive improvement of the
performance, which means that for this KB the information encoded by
these 2 models are complementary. \tatec outperforms all the existing
methods -- except \transe in mean rank -- with a wide margin in
hits@10. \bi\soft performs roughly like \ctransr, and better than its
counterpart \bi\hard. Though \tri\soft is better than \tri\hard as
well, \tatecft\soft and \tatecft\hard converge to very similar
performances. Fine-tuning the parameters is this time better than
simply using a linear combination even if \tateclc is still performing well.

\tatecft outperforms both variants {\sc \tatec-shared} and
{\sc \tatec-no-pretrain} by a wide margin, which confirms that both
pre-training and the use of different embeddings spaces are essential
to properly collect the different data patterns of the \bi and \tri
models: by sharing the embeddings we constrain too much the model, and
without pre-training \tatec is not able to encode the complementary
information of its constituents. The performance of \tatec in these
cases is in-between the performances of the soft version of the \bi
and \tri models, which indicates that they converge to a solution that
is not even able to reach the best performance of their constituent
models.  

We also broke down the results by type of relation, classifying each relationship according to the cardinality of their head and tail arguments. A relationship is considered as 1-to-1, 1-to-M, M-to-1 or M-M regarding the variety of arguments head given a tail and vice versa. If the average number of different heads for the whole set of unique pairs (label, tail) given a relationship is below 1.5 we have considered it as 1, and the same in the other way around. The number of relations classified as 1-to-1, 1-to-M, M-to-1 and M-M is 353, 305, 380 and 307, respectively. The results are displayed in the Table \ref{tab:detailed-res}.  \bi and \tri models cooperate in a constructive way for all the types of relationship when predicting both the head and tail. \tatecft is remarkably better for M-to-M relationships.

\begin{table}[t]
  \caption{\label{tab:detailed-res}{\bf Detailed results by category of relationship.} We compare our {\sc Bigrams}, {\sc Trigram} and \tatec models  in terms of  Hits@10 (in \%) on \fb in the filtered setting against other models of the literature. ({\sc M.} stands for {\sc Many}). }
\begin{center}
\vspace*{-1ex}
\begin{small}
\begin{sc}
\resizebox{1\linewidth}{!}{
\begin{tabular}{|l|cccc|cccc|}
\hline
Task & \multicolumn{4}{c|}{Predicting head} & \multicolumn{4}{c|}{Predicting tail} \\
\hline
Rel. category& 1-to-1 & 1-to-M. & M.-to-1 & M.-to-M. & 1-to-1 & 1-to-M. & M.-to-1 & M.-to-M. \\
\hline
\transe\soft	& 76.2	& 93.6 	& 	47.5 & 70.2	& 76.7  	& 50.9	& 93.1	& 72.9	 	\\
\transh	& 66.8	& 87.6 	& 28.7	& 64.5	& 65.5  	& 39.8 	& 83.3	& 67.2	 	\\
\transr   	& 78.8	& 89.2 	& 34.1	& 69.2	& 79.2 	& 37.4 	& 90.4	& 72.1		\\
\ctransr 	& 81.5 	&  89 &  34.7 & 71.2 & 80.8 	& 38.6	&  90.1 & 73.8	\\
\hline
\bi\soft     &  76.2 	& 90.3 	&  37.4	&  70.1	&  75.9 &  44.4 & 89.8 &  72.8	\\
\tri\soft     &  56.4 	& 79.6 	& 30.2	&  57 	&  53.1 &  28.8 & 81.6 &  60.8	\\
\hdashline
\tatecft\soft     &  79.3	& 93.2 	& 42.3	&  77.2	& 78.5 &  51.5 & 92.7 &   80.7	\\
\hline
\end{tabular}
}
\end{sc}
\end{small}
\end{center}
\vspace*{-4ex}
\end{table}

\subsubsection{\svo}
\tatec achieves also a very good performance on this task since it
outperforms all previous methods on both metrics. As before, both
regularization strategies lead to very similar performances, but the
soft setting is slightly better. In terms of hits@5\%, \tatec
outperforms its constituents, however in terms of mean rank the \bi
model is considerably worse than \tri and \tatec. The performance of
\lfm is in between the \tri and \bi models, which confirms the fact
that sharing the embeddings in the 2- and 3-way terms can actually prevent
to make the best use of both types of interaction.

As for \kinships, since here the performance of \bi is much worse than
that of \tri, \tateclc is very competitive.
It seems that when \bi and \tri perform well for different types of
relationships (such as in \fb), then combining them via fine-tuning
(i.e. \tatecft) allows to get the best of both; however, if one of
them is consistently performing worse on most relationships as it
seems to happen for \kinships and \svo, then \tateclc is a good choice
since it can cancel out any influence of the bad model.
However, Table~\ref{tab:TrainTimes}, depicting training times of various models
on \fb, shows that training \tateclc is around twice as slow as
training \tatecft.

\begin{table}[t]
\caption{{\bf Training times on \fb} on a single core.}
\label{tab:TrainTimes}
\begin{small}
\begin{center}
\begin{sc}
\vspace*{-1ex}
\begin{tabular}{|l|r|}
\hline
Model & Train. time \\
\hline
\bi\soft    & $\sim$ 6h \\
\tri\soft & $\sim$ 12h \\
\tatecft\soft    & $\sim$ 13h \\
\tateclc\soft  & $\sim$22h \\
\hline
\end{tabular}
\vspace*{-4ex}
\end{sc}
\end{center}
\end{small}
\end{table}

\subsection{Illustrative experiments}
This last experimental section provides some illustrations and insights on the performance of \tatec and \transe.

\subsubsection{\transe and symmetrical relationships}

\transe has a peculiar behavior: it performs very well on \fb but quite poorly on all the other datasets.
Looking in detail at \fb, we noticed that this database is made up of a lot of pairs of symmetrical relationships such as \texttt{/film/film/subjects} and \texttt{/film/film\_subject/films}, or  \texttt{/music/album/genre} and \texttt{/music/genre/albums}. The simplicity of the translation model of \transe works well when,
for predicting the validity of an unknown triple, the model can make use of its symmetrical counterpart if it was present in the training set.
Specifically, 45,817 out of 59,071 test triples of \fb have a
symmetrical triple in the training set.  If we split the test triples
into two subsets, one containing the test triples for which a
symmetrical triple has been used in the learning stage and the other
containing those ones for which a symmetrical triple does not exist in the
training set, the overall mean rank of \transe of 50.7 is decomposed
into a mean rank of 17.5 and 165.7, and the overall hits@10 of 71.5
is decomposed into 76.6 and 53.7, respectively.
\transe makes a very adequate use of this particular feature. In the
original \transe paper \cite{bordesNIPS13}, the algorithm is shown to
perform well on \fb and on a dataset extracted from the KB WordNet
\cite{wordnet}: we suspect that the WordNet dataset also contains
symmetrical counterparts of test triples in the training set (such as
hyperonym vs hyponym, meronym vs holonym).

\tatec can also make use of this information and is, as expected,
much better on relations with symmetrical counterparts in train: on
\fb, the mean rank of \tatecft\soft is of 17.5 for relations with
symmetrical counterparts 197.4 instead and hits@10 is of 84.4\% instead of
50\%. Yet, as results on other datasets show, \tatec is also able to
generalize when more complex information needs to be taken into account.

\begin{table}[t]
  \caption{\label{tab:predictionsFB}{\bf Examples of predictions on \fb.} {\small Given an entity and a relation type from a test triple, \tatec fills in the missing slot. In bold is the expected correct answer.}}
\begin{center}
\vspace*{-1ex}
\begin{small}
\begin{tt}
\resizebox{1\linewidth}{!}{
\begin{tabular}{|c|c |}
\hline
{\sc Triple} & {\sc Top-10 predictions} \\
\hline
\multirow{3}{*}{\small{(poland\_national$\_$football$\_$team,  /sports\_team/location, ?)}} & Mexico, South$\_$Africa, \bf{Republic$\_$of$\_$Poland} \\
& Belgium, Puerto$\_$Rico, Austria, Georgia \\
& Uruguay, Colombia, Hong$\_$Kong\\
\hline
\multirow{3}{*}{\small{(?, /film/film$\_$subject/films , remember$\_$the$\_$titans)}} & {\bf racism}, vietnam$\_$war, aviation, capital$\_$punishment \\
& television, filmmaking, Christmas \\
& female, english$\_$language, korean$\_$war\\
\hline
\multirow{3}{*}{\small{(noam$\_$chomsky, /people/person/religion, ?)}} & {\bf atheism}, agnosticism, catholicism, ashkenazi$\_$jews \\
& buddhism, islam, protestantism \\
& baptist, episcopal$\_$church, Hinduism\\
\hline
\multirow{3}{*}{\small{(?, /webpage/category, official$\_$website)}} &  supreme$\_$court$\_$of$\_$canada, butch$\_$hartman, robyn$\_$hitchcoc, mercer$\_$university \\
& clancy$\_$brown, dana$\_$delany, hornets\\
& grambling$\_$state$\_$university, dnipropetrovsk, juanes\\
\hline
\end{tabular}
}
\end{tt}
\end{small}
\end{center}
\vspace*{-4ex}
\end{table}

\subsubsection{Anecdotal examples}
\label{anecdotal}
Some examples of predictions by \tatec on \fb are displayed in
Table~\ref{tab:predictionsFB}. In the first row, we want to know the
answer to the question \texttt{What is the location of the polish
  national football team?}; among the possible answers we find not
only locations, but more specifically countries, which makes sense for
a national team. For the question \texttt{What is the topic of the
  film 'Remember the titans'?} the top-10 candidates may be potential
film topics. Same for the answers to the question \texttt{Which
  religion does Noam Chomsky belong to?} that can all be typed as
religions. In these examples, both sides of the relationship are clearly typed: 
a certain type of entity is expected in head or tail (country, religion, person, movie, etc.). 
The operators of \tatec may then operate on specific regions of the embedding space.
On the contrary, the relationship \texttt{/webpage/category} is an example of non-typed
relationship. This one, which could actually be seen as an attribute
rather than a relationship, indicates if the entity head has a topic
website or an official website. Since many types of entities can have
a webpage and there is little to no correlation among relationships,
predicting the left-hand side argument is nearly impossible.

\begin{figure}
\hspace*{-0.8cm}
\begin{subfigure}{0.55\textwidth}
\includegraphics[width=\linewidth]{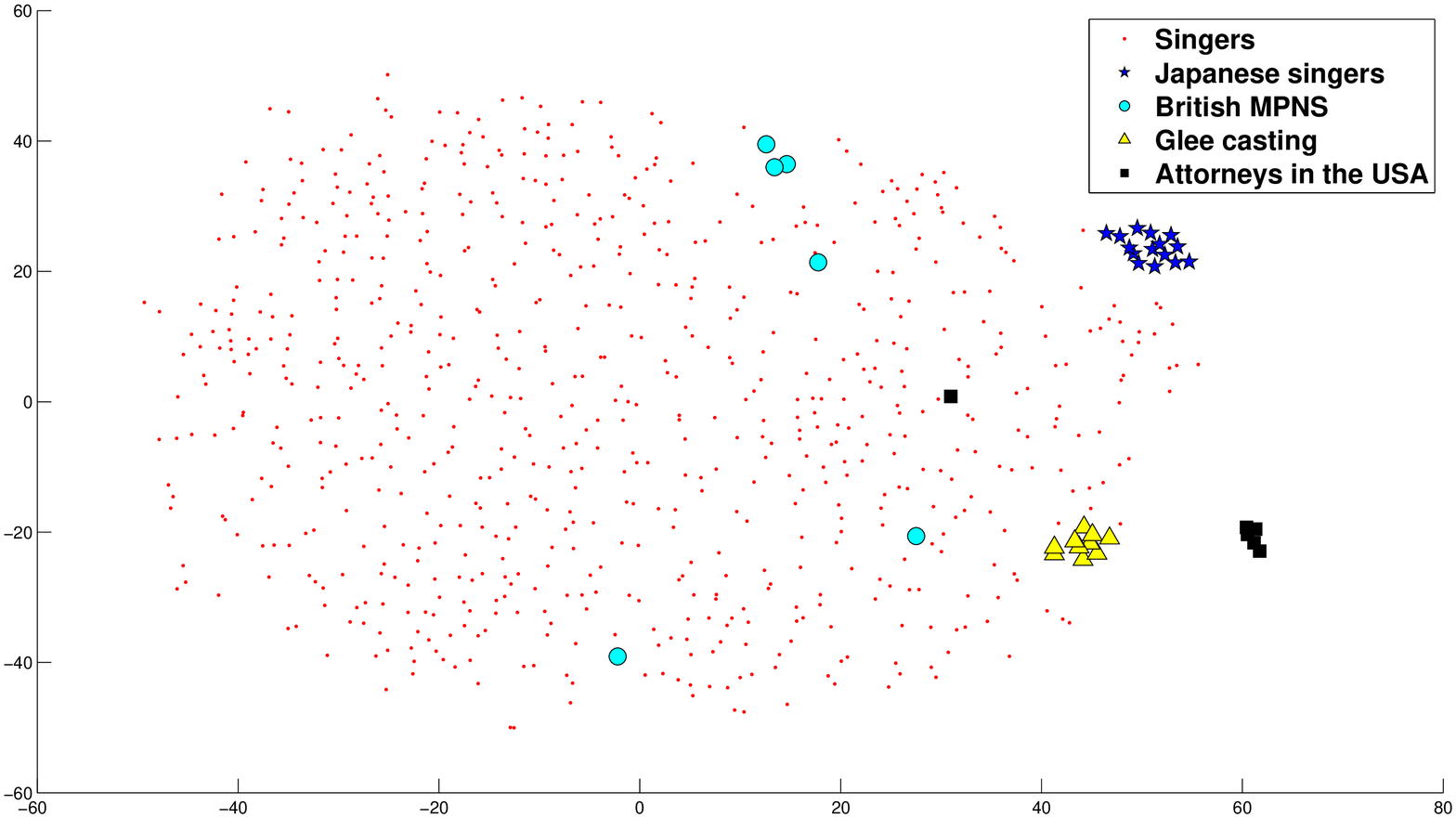}
\caption{Embeddings of \tri} \label{fig:2a}
\end{subfigure}
\hspace*{-0.5cm} 
\begin{subfigure}{0.55\textwidth}
\includegraphics[width=\linewidth]{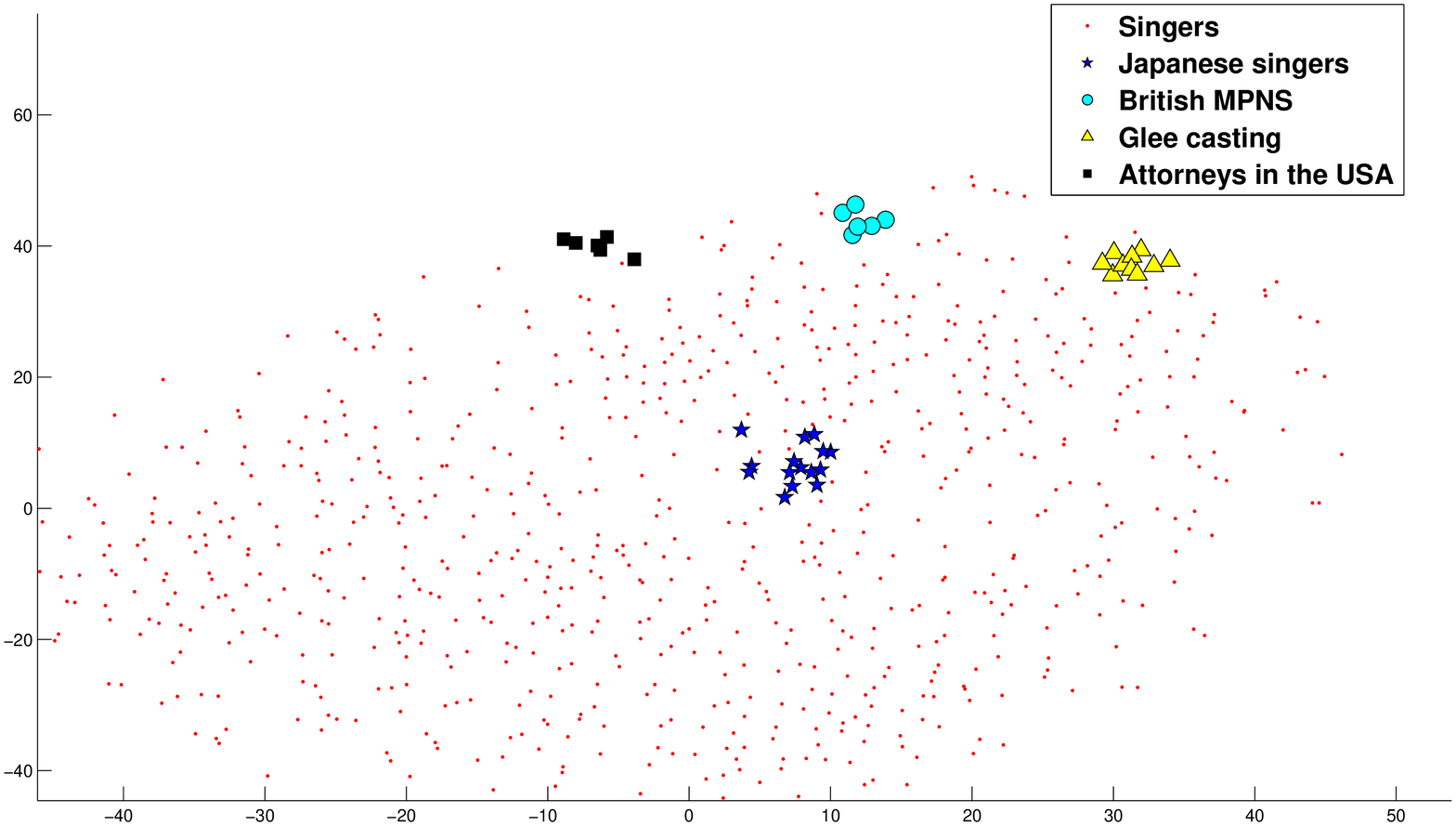}
\caption{Embeddings of \bi} \label{fig:2b}
\end{subfigure}
\caption{{\bf Embeddings obtained by \tri
  and \bi models} and projected in 2-D using t-SNE. MPNS stands for \texttt{Main Profession is Not Singer}.}
\end{figure}

Figures \ref{fig:2a} and \ref{fig:2b} show 2D projections of
embeddings of selected entities for the \tri and \bi models trained on
\fb, respectively, obtained by projecting them using t-SNE
\cite{van2008visualizing}. This projection has been carried out only
for Freebase entities whose profession is either {\tt singer} or {\tt attorney} in
the {\tt USA}. We can observe in Figure \ref{fig:2a} that all attorneys are
clustered and separated from the singers, except one, which
corresponds to the multifaceted {\tt Fred Thompson}\footnote{Apart
  from being an attorney, he is an actor, a radio
  personality, a lawyer and a politician}. However, embeddings of the
singers are not clearly clustered: since singers can appear in a
multitude of triples, their layout is the result of a compendium of
(sometimes heterogeneous) categories. To illustrate graphically the
different data patterns to which \bi and \tri respond, we focus on the
small cluster made up of Japanese singers that can be seen in Figure
\ref{fig:2a} (\tri). In Figure \ref{fig:2b} (\bi) however, these same
entities are more diluted in the whole set of singers. Looking at the
neighboring embeddings of these Japanese singers entities in Figure
\ref{fig:2b}, we find entities highly connected to {\tt japan} like
{\tt yoko\_ono}
-- born in Japan, {\tt vic\_mignogna}, {\tt greg\_ayres}, {\tt
  chris\_patton} or {\tt laura\_bailey} -- all of them worked in the dubbing industry of Japanese
\textit{anime} movies and television series.
This shows the impact of the interaction between heads and tails
in the \bi model: it tends to push together entities connected in
triples whatever the relation. In this case, this forms a Japanese cluster.

\begin{table}[t]
  \caption{\label{tab:predictionsSVO}{\bf Examples of predictions on \svo.} {\small  Given two nouns acting as subject and direct object from a test triple, \tatec predicts the best fitting verb.  In bold is the expected correct answer.}}
\begin{center}
\vspace*{-1ex}
\begin{small}
\begin{tt}
\resizebox{0.8\linewidth}{!}{
\begin{tabular}{|c|c |}
\hline
{\sc Triple} & {\sc Top-10 predictions} \\
\hline
\multirow{2}{*}{\small{(bus, ? , service)}} & use, provide, {\bf run}, have, include \\
& carry, offer, enter, make, take \\
\hline
\multirow{2}{*}{\small{(emigrant, ? , country)}} & flee, become, {\bf enter}, leave, form \\
& dominate, establish, make, move, join \\
\hline
\multirow{2}{*}{\small{(minister, ?, protest)}} & lead, organize, {\bf join}, involve, make \\
& participate, conduct, stag, begin, attend \\
\hline
\multirow{2}{*}{\small{(vessel, ?, coal)}} & use, {\bf transport}, carry, convert, send \\
& make, provide, supply, sell, contain \\
\hline
\multirow{2}{*}{\small{(tv$\_$channel, ?, video)}} & feature, make, release, use, produce \\
& have, include, call, base, show \\
\hline
\multirow{2}{*}{\small{(great$\_$britain, ?, north$\_$america)}} & include, become, found, establish, dominate \\
& name, have, enter, form, run \\
\hline
\end{tabular}
}
\end{tt}
\end{small}
\end{center}
\vspace*{-4ex}
\end{table}

Table \ref{tab:predictionsSVO} shows examples of predictions on
\svo. In the first example, though \texttt{run} is the target verb for
the pair \texttt{(bus, service)}, other verbs like \texttt{provide} or
\texttt{offer} are good matches as well. Similarly, non-target verbs
like \texttt{establish} or \texttt{join}, and {\tt lead}, {\tt
  participate} or {\tt attend} are good matches for the second and
third examples (\texttt{(emigrant, country)} and \texttt{(minister,
  protest)}) respectively. The fourth and fifth instances show an
example of very heterogeneous performance for a same relationship (the
target verb is {\tt transport} in both cases) which can be easily explained from a
semantic point of view: transport is a very good fit given the pair
(\texttt{vessel}, \texttt{coal}), whereas \texttt{a TV channel
  transports video} is not a very natural way to express that one can
watch videos in a TV channel, and hence this leads to a very poor
performance -- the target verb is ranked \#696. The sixth
example is particularly interesting, since even if the target verb,
\texttt{colonize}, is ranked very far in the list (\#344), good
candidates for the pair \texttt{(Great Britain, North America)} can be
found in the top-10. Some of them have a similar representation as
\texttt{colonize}, because they are almost synonyms, but they
are ranked much higher. This is an effect of the verb
frequency.

\begin{figure}
\begin{subfigure}{0.49\textwidth}
\includegraphics[width=1.05\linewidth]{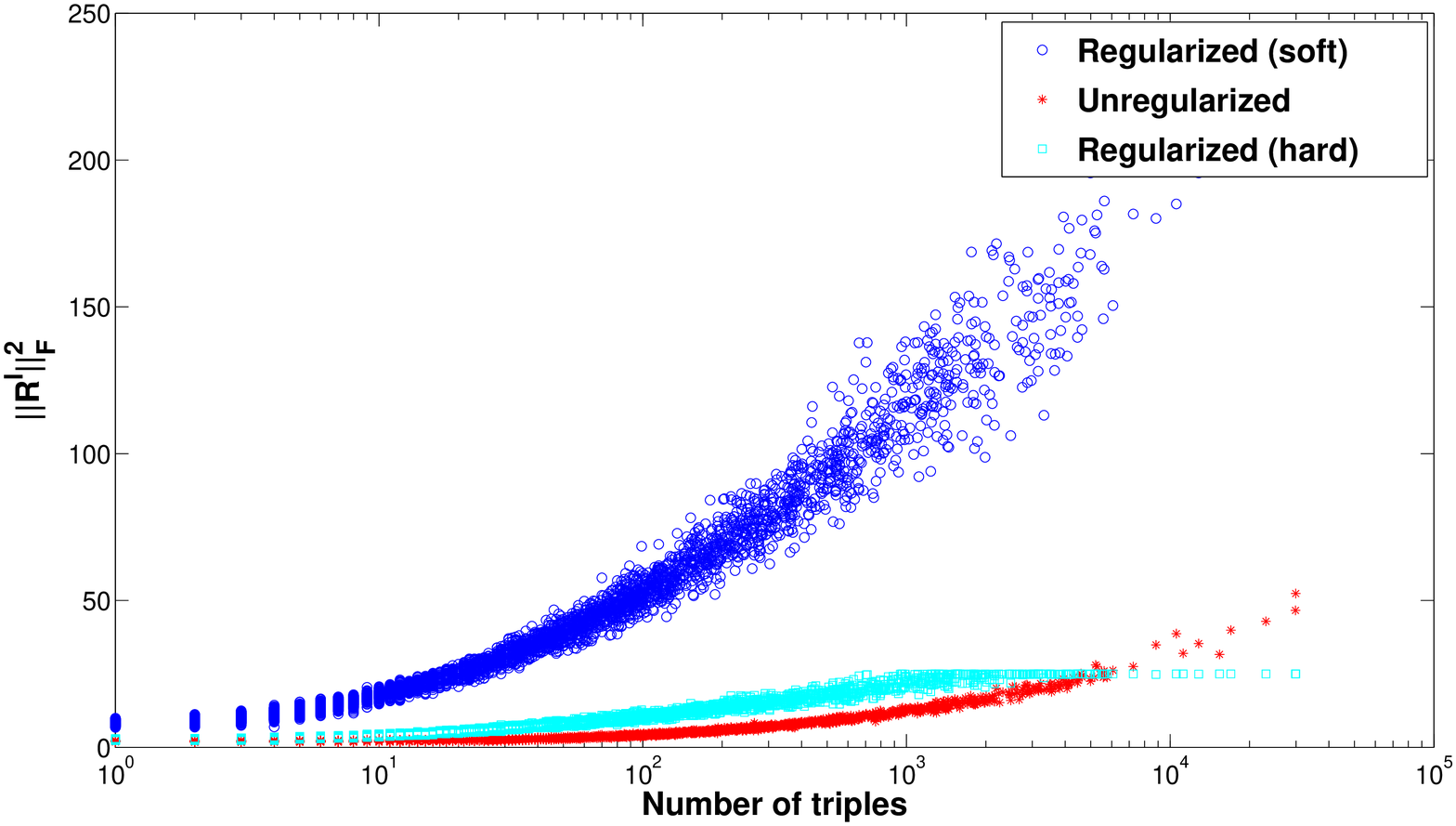}
\caption{Frobenius norm of the rel. matrices according to the number
  of training triples of each rel.} \label{fig:1a}
\end{subfigure}
\hspace*{\fill} 
\begin{subfigure}{0.49\textwidth}
\includegraphics[width=\linewidth]{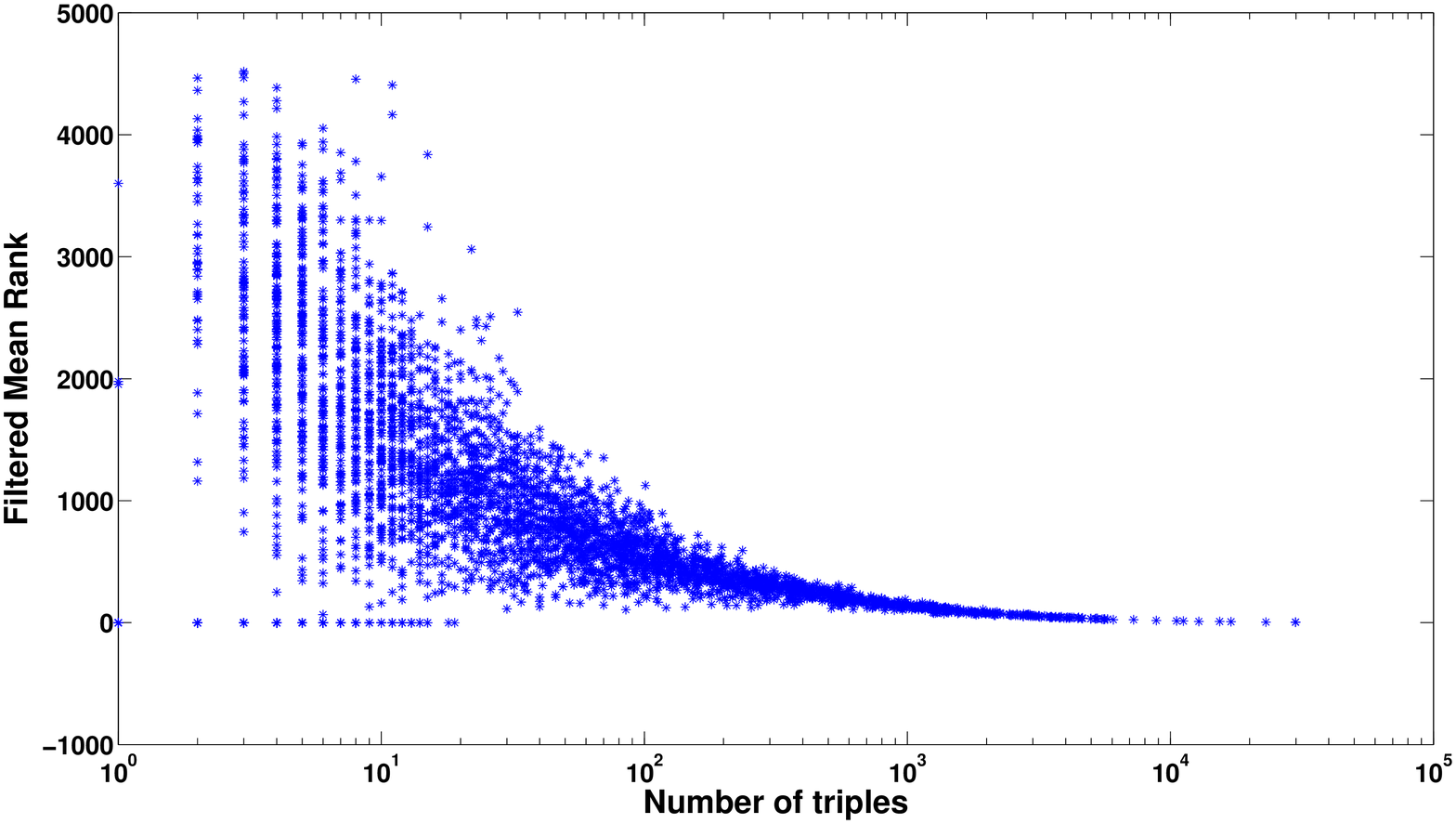}
\caption{Test mean rank according to the number of training triples of each relationship.} \label{fig:1b}
\end{subfigure}
\caption{Indicators of the behavior of \tatecft on \fb according to to the number of training triples of each relationship.}

\end{figure}

As illustrated in Figure \ref{fig:1a}, the more frequent a
relationship is, the higher its Frobenius norm is; hence, verbs with
similar meanings but unbalanced frequencies can be ranked differently,
which explains that a rare verb, such as {\tt colonize}, can be ranked
much worse than other semantically similar words. A consequence of
this relation between the Frobenius norm and the appearance frequency
is that usual verbs tend to be highly ranked even though sometimes
they are not good matches, due to the influence of the norm in the
score. In that figure, we can see that the Frobenius norm of the
relation matrices are larger in the regularized (\textit{soft}) case
than in the unregularized case. This happens because we fixed a very
large value for both $C_2$ and $\rho_l$ in the regularized case
($\rho_e$ is fixed to 1). It imposes a strong constraint on the norm
of the entities but not on the relationship matrices and makes the
Frobenius norm of these matrices absorb the whole impact of the norm
of the score, and, thus, the impact of the verb frequency. We could
down-weight the importance of the verb frequency by tuning the
parameters $\rho_l$ and $C_2$ to enforce a stronger constraint. Figure
\ref{tab:predRegVSUnreg} shows the effect of the verb frequency in
these two models when predicting the same missing verb as in
Table~\ref{tab:predictionsSVO}.

\begin{table}[t]
  \caption{\label{tab:predRegVSUnreg}{\bf Examples of predictions on \svo for a regularized and an unregularized \tri.} {\small In bold is the expected correct answer.}}
\begin{center}
\vspace*{-1ex}
\begin{small}
\begin{tt}
\resizebox{1\linewidth}{!}{
\begin{tabular}{|c|c|c|}
\hline
\multirow{2}{*}{\sc Triple} & \multicolumn{2}{c|}{\sc Top-10 predictions} \\
\cline{2-3}
 & Unregularized & Regularized (soft) \\
\hline
\multirow{2}{*}{\small{(bus, ? , service)}} & use, operate, offer, call, build, &  provide, use, have, include, make,\\
& include, have, know, make, create & offer, take, carry, serve, {\bf run}\\ 
\hline
\multirow{2}{*}{\small{(emigrant, ? , country)}} & use, represent, save, flee, visit, & flee, become, come, {\bf enter}, found,\\
&come, make, leave, create, know & include, form, make, leave, join\\ 
\hline
\multirow{2}{*}{\small{(minister, ? , protest)}} & bring, lead, reach, have, become, &  lead, organize, conduct,  participate, {\bf join}\\
& say, include, help, leave, appoint & make, involve, support, suppress, raise\\ 
\hline
\multirow{2}{*}{\small{(vessel, ? , coal)}} & take, use, have, carry, make,   &  use, {\bf transport}, make, carry, deliver,  \\
& hold, move, become, fill, serve & send, contain, supply, leave, provide\\ 
\hline
\multirow{2}{*}{\small{(tv$\_$channel, ?, video)}} & make, include, write, know, have,  &  release,  make, feature, produce, have, \\
& produce, use, play, give, become & include, use, take, show, base\\ 
\hline
\multirow{2}{*}{\small{(great$\_$britain, ?, north$\_$america)}} & have, use, include, make, leave, & include, found, become, run,name, \\
& become, know, take, call, build & move, annex, form, establish, dominate\\ 
\hline
\end{tabular}
}
\end{tt}
\end{small}
\end{center}
\vspace*{-4ex}
\end{table}


Breaking down the performance by relationship, this is translated into
a strong relation between the performance of a relationship and its
frequency (see Figure \ref{fig:1b}). However, the same relation
between the $2$-norm of the entities embeddings and their frequency is
not observed, which can be explained given that an entity can appear
in the left and right argument in an unbalanced way.
%

\section{Conclusion}
This paper presents \tatec, a tensor factorization method
that satisfactorily combines 2- and 3-way interaction terms to obtain
a performance better than the best of either constituent. Different
data patterns are properly encoded thanks to the use of different
embedding spaces and of a two-phase training (pre-training and
fine-tuning/linear-combination).  Experiments on four benchmarks for
different tasks and with different quality measures prove the strength
and versatility of this model, which could actually be seen as a
generalization of a lot of existing works. Our experiments also allow
us to draw some conclusions about the two usual regularization schemes
used so far in these embedding-based models: they both achieve similar
performances, even if soft regularization appears slightly more
efficient but with one extra-hyperparameter.

\acks{This work was carried out in the framework of the Labex MS2T (ANR-11-IDEX-0004-02), and was funded by the French National Agency for Research (EVEREST-12-JS02-005-01).}

\appendix
\appendixpage
\section{Optimal hyperparameters}
\label{app:hp}
The optimal configurations for \umls are:\\
-~\transe\soft: $d=40, \lambda=0.01, \gamma = 0.5, C=0$; \\
-~\bi\soft: $d_1=40, \lambda_1=0.01, \gamma = 0.5, C=0.1$; \\
-~\tri\soft: $d_2=40, \lambda_2=0.01, \gamma = 1, C=0.1, \rho_l=5$; \\
-~\tatec\soft: $d_1=40, d_2=40, \lambda_1=\lambda_2=0.001, \gamma = 1, C_1= C_2=0.01, \rho_l=5$; \\
-~\transe\hard: $d=40, \lambda=0.01, \gamma = 0.1$; \\
-~\bi\hard: $d_1=40, \lambda_1=0.01, \gamma = 0.5$; \\
-~\tri\hard: $d_2=40, \lambda_2=0.01, \gamma = 1, \rho_l=10$; \\
-~\tatec\hard: $d_1=40, d_2=40, \lambda_1=\lambda_2=0.001, \gamma = 1, \rho_l=10$.\\
-~{\sc \tatec-linear-comb}: $d_1=40, d_2=40, \gamma = 0.5,  \alpha=50$.\\

\noindent
The optimal configurations for \kinships are:\\
-~\transe\soft: $d=40, \lambda=0.01, \gamma = 1, C=0$;\\
-~\bi\soft: $d_1=40, \lambda_1=0.01, \gamma = 1, C=1$;\\
-~\tri\soft: $d_2=40, \lambda_2=0.01, \gamma = 0.5, C=0.1, \rho_l=5$;\\
-~\tatec\soft: $d_1=40, d_2=40, \lambda_1=\lambda_2=0.001, \gamma = 1, C_1=100, C_2=0.0001, \rho_l=10$; \\
-~\transe\hard: $d=40, \lambda=0.01, \gamma = 1$;\\
-~\bi\hard: $d_1=40, \lambda_1=0.01,\gamma = 1$;\\
-~\tri\hard: $d_2=40, \lambda_2=0.01, \gamma = 0.5, \rho_l=10$;\\
-~\tatec\hard $d_1=40, d_2=40, \lambda_1=\lambda_2=0.001, \gamma = 1, \rho_l=10$.\\
-~{\sc \tatec-linear-comb}: $d_1=40, d_2=40, \gamma = 1,  \alpha=10$.\\

\noindent
The optimal configurations for \fb are: \\
-~\transe\soft: $d=100, \lambda=0.01, \gamma = 0.25, C=0.1$;\\
-~\bi\soft: $d_1=100, \lambda_1=0.01, \gamma = 1, C=0$;\\
-~\tri\soft: $d_2=50, \lambda_2=0.01, \gamma = 0.25, C=0.001, \rho_l=1$;\\
-~\tatec\soft: $d_1=100, d_2=50, \lambda_1=\lambda_2=0.001,  \gamma = 0.5, C_1=C_2=0$;\\
-~\transe\hard: $d=100, \lambda=0.01, \gamma = 0.25$;\\
-~\bi\hard: $d_1=100, \lambda_1=0.01, \gamma = 0.25$;\\
-~\tri\hard: $d_2=50, \lambda_2=0.01, \gamma = 0.25, \rho_l=5$;\\
-~\tatec\hard: $d_1=100, d_2=50, \lambda_1=\lambda_2=0.001,  \gamma = 0.25, \rho_l=5$;\\
-~{\sc \tatec-no-pret}: $d_1=100, d_2=50, \lambda_1=\lambda_2=0.01,  \gamma = 0.25,  C_1=0, C_2=0.001, \rho_l=1$;\\
-~{\sc \tatec-shared}: $d_1=d_2=75, \lambda_1=\lambda_2=0.01,  \gamma = 0.25,  C_1=C_2=0.001, \rho_l=5$;\\
-~{\sc \tatec-linear-comb}: $d_1=100, d_2=50, \gamma = 0.25,  \alpha=200$.\\

\noindent
The optimal configurations for \svo are: \\
-~\transe\soft: $d=50, \lambda=0.01, \gamma = 0.5, C=1$;\\
-~\bi\soft: $d_1=50, \lambda_1=0.01, \gamma = 1, C=0.1$;\\
-~\tri\soft: $d_2=50, \lambda_2=0.01, \gamma = 1, , C=10, \rho_l=20$;\\
-~\tatec\soft: $d_1=50, d_2=50, \lambda_1=\lambda_2=0.0001,  \gamma = 1, C_1=0.1, C_2=1, \rho_l=20$;\\
-~\transe\hard: $d=50, \lambda=0.01, \gamma = 0.5$;\\
-~\bi\hard: $d_1=50, \lambda_1=0.01, \gamma = 1$;\\
-~\tri\hard: $d_2=50, \lambda_2=0.01, \gamma = 1, \rho_l=20$;\\
-~\tatec\hard: $d_1=50, d_2=50, \lambda_1=\lambda_2=0.0001, \gamma = 1, \rho_l=20$.\\
-~{\sc \tatec-linear-comb}: $d_1=50, d_2=50, \gamma = 1,  \alpha=50$.

\bibliography{jair15}
\bibliographystyle{theapa}

\end{document}